\documentclass[conference]{IEEEtran}
\usepackage{url}
\IEEEoverridecommandlockouts
\usepackage{cite}
\usepackage{amsmath,amssymb,amsfonts}
\usepackage{algorithmic}
\usepackage{graphicx}
\usepackage{textcomp}
\usepackage{xcolor}
\usepackage{tikz}
\usetikzlibrary{shapes,positioning, shapes.multipart}
\usepackage{pgfplots}
\pgfplotsset{compat=1.18}
\usepackage{booktabs}

\def\BibTeX{{\rm B\kern-.05em{\sc i\kern-.025em b}\kern-.08em
    T\kern-.1667em\lower.7ex\hbox{E}\kern-.125emX}}
\begin{document}

\title{Efficient Neural Architectures for Real-Time ECG Interpretation on Limited Hardware%
\thanks{© 2025 IEEE. Personal use of this material is permitted.
Permission from IEEE must be obtained for all other uses, in any current
or future media, including reprinting/republishing this material for
advertising or promotional purposes, creating new collective works, for
resale or redistribution to servers or lists, or reuse of any copyrighted
component of this work in other works.
Published in: 2025 IEEE International Conference on Big Data (BigData).
DOI: 10.1109/BIGDATA66926.2025.11402097}%
}

\author{\IEEEauthorblockN{Ashery Mbilinyi}
\IEEEauthorblockA{\textit{Department of Computer Science} \\
\textit{University of Victoria}\\
Victoria, Canada \\
ashery@uvic.ca}
\and
\IEEEauthorblockN{Callum O'Riley}

\IEEEauthorblockA{\textit{Department of Electrical and Computer Engineering} \\
\textit{University of British Columbia}\\
Vancouver, Canada \\
callumo@student.ubc.ca}
\and
\IEEEauthorblockN{Julia Handra}
\IEEEauthorblockA{\textit{Faculty of Medicine} \\
\textit{University of British Columbia}\\
Vancouver, Canada \\
jhandra@student.ubc.ca}
\and
\IEEEauthorblockN{Ashley Moller-Hansen}
\IEEEauthorblockA{\textit{School of Biomedical Engineering} \\
\textit{University of British Columbia}\\
Vancouver, Canada \\
ashley.mollerhansen@ubc.ca}
\and
\IEEEauthorblockN{Jason Andrade}
\IEEEauthorblockA{\textit{Division of Cardiology}\\
\textit{Faculty of Medicine}\\
\textit{University of British Columbia}\\
Vancouver, Canada \\
jason.andrade@vch.ca}
\and
\IEEEauthorblockN{Marc Deyell}
\IEEEauthorblockA{\textit{Division of Cardiology} \\
\textit{Faculty of Medicine}\\
\textit{University of British Columbia}\\
Vancouver, Canada \\
mdeyell@providencehealth.bc.ca}
\and
\IEEEauthorblockN{Cameron Hague}
\IEEEauthorblockA{\textit{Department of Radiology} \\
\textit{Faculty of Medicine}\\
\textit{University of British Columbia}\\
Vancouver, Canada \\
cjhague75@gmail.com}
\and
\IEEEauthorblockN{Nathaniel Hawkins}
\IEEEauthorblockA{\textit{Division of Cardiology} \\
\textit{Faculty of Medicine}\\
\textit{University of British Columbia}\\
Vancouver, Canada \\
nhawkins@mail.ubc.ca}
\and
\IEEEauthorblockN{Kendall Ho}
\IEEEauthorblockA{\textit{Department of Emergency Medicine} \\
\textit{Faculty of Medicine}\\
\textit{University of British Columbia}\\
Vancouver, Canada \\
kendall.ho@ubc.ca}
\and
\IEEEauthorblockN{Jonathan Leipsic}
\IEEEauthorblockA{\textit{Department of Radiology} \\
\textit{Faculty of Medicine}\\
\textit{University of British Columbia}\\
Vancouver, Canada \\
jonathan.leipsic@ubc.ca}
\and
\IEEEauthorblockN{Roger Tam}
\IEEEauthorblockA{\textit{School of Biomedical Engineering} \\
\textit{University of British Columbia}\\
Vancouver, Canada \\
roger.tam@ubc.ca}
}

\maketitle

\begin{abstract}
Electrocardiogram (ECG) interpretation is essential for diagnosing a wide range of cardiac abnormalities. While deep learning has shown strong potential for automating ECG classification, many existing models rely on large, computationally intensive architectures that hinder practical deployment. In this paper, we present an empirical study of convolutional neural network (CNN) architectures, exploring tradeoffs between diagnostic accuracy and computational efficiency. We benchmark two established baselines: AttiaNet, a compact model composed of sequential temporal and spatial blocks, and DeepResidualCNN, the winning architecture of the 2021 PhysioNet/Computing in Cardiology Challenge. Building on these, we propose three lightweight models: (i) ParallelCNN, which employs dual temporal and spatial branches for parallel pattern extraction; (ii) ParallelCNNew, a variant with symmetric weight initialization for balanced feature learning; and (iii) SimpleNet, a streamlined architecture that jointly processes temporal and spatial dimensions. Our experiments span three publicly available 12-lead ECG datasets from Germany, China, and the United States, covering binary, multiclass, and multilabel classification tasks across diverse patient populations. We further evaluate the impact of integrating low-cost demographic metadata (age and sex) to improve performance with minimal overhead. To ensure fair comparison, we introduce a unified \textit{Efficiency Score} that integrates model size, inference speed, memory usage, and AUC performance. By balancing diagnostic performance and efficiency, our models offer a scalable and viable foundation for next-generation AI systems in cardiovascular care.
\end{abstract}

\begin{IEEEkeywords}
Convolutional neural networks, electrocardiography, medical signal processing, model efficiency, real-time systems, resource-constrained systems, spatiotemporal data analysis.
\end{IEEEkeywords}

\section{Introduction}
\label{sec:introduction}
Cardiovascular disease (CVD) remains the leading cause of mortality in both the United States \cite{ahmad2024leading} and worldwide \cite{WHO2024, handra2024role, ribeiro2020automatic}, highlighting the urgent need for reliable tools that enable early detection and intervention. Electrocardiography (ECG) is one of the most widely adopted diagnostic techniques for assessing cardiac health, offering a non-invasive and cost-effective means of identifying conditions such as myocardial infarction, arrhythmias, ischemic heart disease, and cardiomyopathies \cite{ansari2023deep}. Despite its clinical value, ECG interpretation is cognitively demanding and highly dependent on the expertise of specialists in cardiac electrophysiology and waveform analysis \cite{cook2020accuracy, wood2014exploring}, making it vulnerable to diagnostic variability and human error \cite{goy2013competency, anh2006accuracy}.

Deep learning, particularly convolutional neural networks (CNNs), has emerged as a promising approach to automate ECG interpretation by learning complex temporal and morphological patterns directly from raw signals \cite{rafie2021ecg, petmezas2022state, kolk2023machine, herman2024validation, ansari2023deep, ribeiro2020automatic}. While these models achieve impressive diagnostic accuracy, state-of-the-art architectures often involve millions of parameters and high computational demands, resulting in long inference times and substantial memory requirements. Consequently, their deployment is largely confined to settings with high-performance computing resources, limiting their utility in resource-constrained hospitals, mobile clinics, or point-of-care devices where real-time interpretation is most critical \cite{jia2023importance}.

This growing gap between algorithmic advances and deployment feasibility emphasizes the need for neural architectures that balance diagnostic accuracy with computational demand. To address this challenge, we present an empirical study of five CNN-based models for ECG interpretation, explicitly designed and evaluated under limited hardware constraints. Our study includes two established baselines---a compact network and a deep residual CNN from the literature---alongside three lightweight architectures developed from first principles with efficiency as the primary design objective. These models are evaluated on three large and diverse 12-lead ECG datasets from Germany, China, and the United States, covering binary, multiclass, and multilabel classification tasks. Beyond efficiency-focused modeling, we also examine whether integrating low-cost demographic information (age and sex) can improve diagnostic accuracy without introducing significant computational overhead, given their routine availability and clinical relevance.

The key contributions of this work are as follows:

\begin{itemize}
    \item We propose three novel models optimized for efficient ECG interpretation and compare them with compact and deep CNN baselines.
    \item We evaluate all models on three diverse 12-lead ECG datasets across binary, multiclass, and multilabel classification tasks, ensuring robustness and generalizability.
    \item We examine the impact of incorporating age and sex, showing that clinically relevant metadata can enhance performance at negligible computational cost.
    \item To enable fair comparison, we propose a unified efficiency evaluation metric that integrates model size, inference speed, memory usage, and diagnostic performance.
\end{itemize}

The remainder of this paper is organized as follows: Section~\ref{related_work} reviews related work on ECG interpretation and efficient neural networks; Section~\ref{model_architectures} outlines both the relevant literature and our proposed lightweight architectures; Section~\ref{sec:datasets} describes the datasets and classification tasks used for evaluation; Section~\ref{experiments_and_results} reports the experimental results; Section~\ref{discussion} discusses key findings, limitations, and clinical relevance; and Section~\ref{conclusion} concludes with final remarks and directions for future research.

\begin{figure*}[ht]
  \centering
  \includegraphics[width=0.7\linewidth]{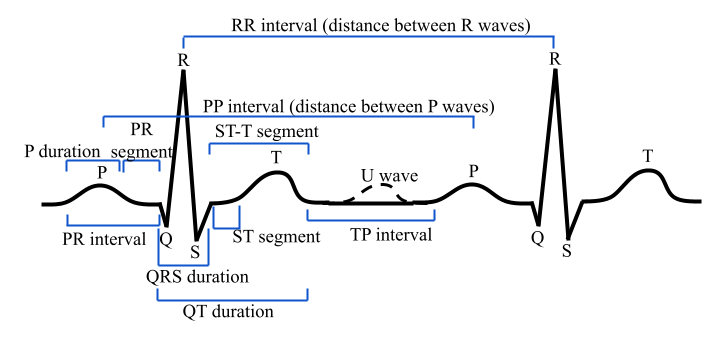}
  \caption{The standard ECG curve with its most common waveforms. Important intervals and points of measurement are depicted. Manual ECG interpretation requires knowledge of these waves and intervals~\cite{clinicalecginterpretation}.}
  \label{fig:ecg}
\end{figure*}

\section{Related Work}
\label{related_work}
\subsection{Computation-Efficient Neural Architectures}

The development of computationally efficient AI models for healthcare has become increasingly important for real-world deployment, particularly in environments lacking access to high-end infrastructure. Lightweight architectures not only reduce energy consumption and latency, but also enable real-time inference on edge devices and in low-resource clinical settings. Recent studies have highlighted the widening gap between the growing computational demands of modern deep learning models and the limited infrastructure available in practice. For example, Jia et al.~\cite{jia2023importance} emphasize the unsustainable growth in model complexity---ranging from 7 million parameters in U-Net~\cite{ronneberger2015u} to more than 2 billion in the Swin Transformer~\cite{wei2023high}---and argue for resource-conscious model design.

To mitigate these challenges, researchers have explored pruning, quantization, model compression, and neural architecture search (NAS). While these methods can yield lighter and faster models, they often introduce substantial computational overhead. For example, applying NAS to transformer models such as BERT can require up to six times the training cost of the original model~\cite{strubell2020energy}.

In the ECG domain, Zhang et al.~\cite{zhang2022rt} applied NAS for efficient signal reconstruction, while others have explored fairness-aware pruning~\cite{wu2022fairprune} and quantization for edge devices~\cite{xu2018efficient, zhang2021medq}. Despite these advances, such models are frequently post hoc adaptations of large networks, which limits their deployability in resource-constrained environments.

In contrast, our work adopts a straightforward design strategy: rather than compressing existing models, we develop three task-specific CNNs---SimpleNet, ParallelCNN, and ParallelCNNew---from the ground up with computational efficiency as the primary objective. These architectures are designed to minimize parameter count and inference time while preserving strong diagnostic accuracy, achieving a practical balance between performance and deployability.

\subsection{Efficient ECG Interpretation}

Parallel to the development of general efficiency techniques, another stream of research has focused on large-scale foundational models for ECG interpretation. These models, such as those proposed by Vaid et al.~\cite{vaid2023foundational} and McKeen et al.~\cite{mckeen2024ecg}, take advantage of massive pre-training to learn broadly generalizable ECG representations. Although powerful, they come at a significant computational cost. For example, McKeen et al.'s model contains 311 million parameters and required two weeks of training on four NVIDIA A100 80GB GPUs---resources far beyond what is practical for most clinical institutions.

Our study explores an alternative path by emphasizing compact, task-specific CNNs that are explicitly optimized for both diagnostic accuracy and computational efficiency. For example, our smallest model, SimpleNet, contains just 2.59 million parameters but delivers rapid inference across various classification tasks. Conceptually, our work is related to Rahhal et al.~\cite{al2018convolutional}, who applied CNNs for arrhythmia detection using transfer learning from ImageNet. However, we instead design ECG-specific architectures from first principles and rigorously evaluate them in binary, multiclass, and multilabel classification tasks.

Furthermore, we extend previous work by evaluating performance on three diverse, publicly available 12-lead ECG datasets: PTB-XL (Germany), Chapman-Shaoxing (China), and MIMIC-IV (USA), spanning different clinical contexts and patient populations. Finally, unlike most efficiency-focused studies, we examine the role of low-cost demographic metadata (age and sex) in improving diagnostic performance. These features are clinically relevant and computationally inexpensive to integrate, making them especially valuable for real-time systems deployed under limited hardware constraints.

\section{Model Architectures}
\label{model_architectures}

Our model design and selection were guided by two key considerations. First, ECG signals can be represented as structured two-dimensional matrices, where one axis corresponds to the 12 spatially distributed leads and the other captures temporal information. For example, a standard 12-lead ECG sampled at 500 Hz over 10 seconds produces a \(12 \times 5000\) matrix. In this representation, each lead provides a unique spatial perspective of cardiac activity, while the temporal axis captures waveform evolution---including the P wave, QRS complex, T wave, and occasionally the U wave (see Fig.~\ref{fig:ecg}). Second, clinical ECG interpretation relies on recognizing both waveform morphology and timing intervals. Cardiologists analyze shape variations and temporal sequences to detect abnormalities such as arrhythmias, myocardial infarction, or conduction disorders. This dual emphasis on spatial and temporal features motivates the use of CNNs, whose kernels can be tailored and receptive fields adjusted to capture such patterns.

Based on these principles, we evaluated five CNN architectures for ECG classification. Two models serve as baselines: AttiaNet~\cite{attia2019screening}, a compact network with sequential temporal and spatial blocks, and DeepResidualCNN~\cite{nejedly2021classification}, a deep residual architecture that achieved state-of-the-art performance in the 2021 PhysioNet/Computing in Cardiology Challenge~\cite{reyna2021will}. In addition, we introduce three
lightweight architectures---ParallelCNN, ParallelCNNew, and SimpleNet---designed from the ground up with computational efficiency as a primary objective. These models vary in depth, parameter count, and convolutional block design, allowing
a systematic investigation of spatiotemporal feature extraction while examining trade-offs between diagnostic accuracy and computational demands.

Below, we provide detailed descriptions of each model.

\subsection{DeepResidualCNN}
\label{subsec:deepresidualcnn}
The DeepResidualCNN was the winning solution in the PhysioNet/Computing in Cardiology Challenge 2021~\cite{nejedly2021classification} and is inspired by the ResNet family of architectures~\cite{He_2016_CVPR}. Its design begins with a 2D convolutional layer that transforms the \(12 \times T\) input into 256 feature maps using a kernel size of \(1 \times 15\) and a stride of \(1 \times 2\), followed by batch normalization and a LeakyReLU activation. The core of the model consists of five residual blocks, each containing two convolutional layers with a kernel size of \(1 \times 9\), batch normalization, and LeakyReLU activation. Downsampling, when required, is handled by an identity path that combines average pooling with a \(1 \times 1\) convolution. After the residual stack, dropout (\(p = 0.5\)) is applied, and adaptive max pooling reduces the temporal dimension to produce a 256-dimensional vector representation. This representation is then passed through both a main classifier and an auxiliary head, each composed of two fully connected layers with batch normalization and LeakyReLU activation.

\subsection{AttiaNet}
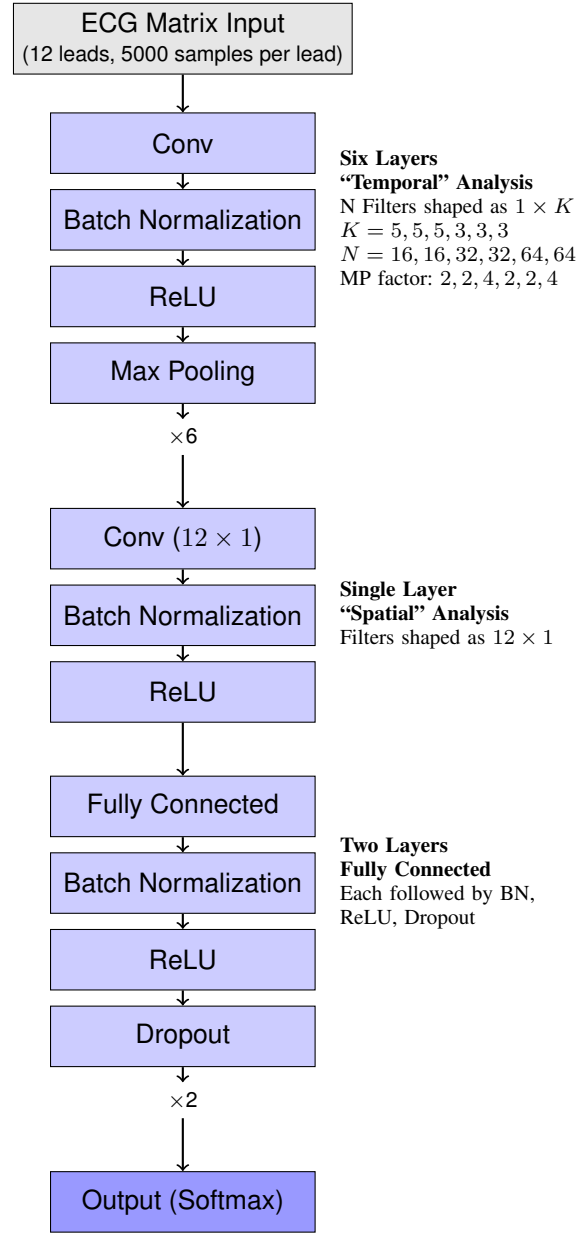
\begin{figure}[ht]
\centering
\begin{tikzpicture}[
    block/.style={rectangle, draw=black, fill=blue!20, minimum width=3.5cm, minimum height=0.8cm, text centered},
    smallblock/.style={rectangle, draw=black, fill=blue!10, minimum width=3.5cm, minimum height=0.8cm, text centered},
    connector/.style={->, thick},
    label/.style={text width=6cm, align=left, font=\footnotesize},
    font=\sffamily
]

\node[block, fill=gray!20] (input) {\shortstack{ECG Matrix Input\\ \footnotesize (12 leads, 5000 samples per lead)}};

\node[block, below=0.5cm of input] (conv1) {Conv};
\node[block, below=0.2cm of conv1] (bn1) {Batch Normalization};
\node[block, below=0.2cm of bn1] (relu1) {ReLU};
\node[block, below=0.2cm of relu1] (pool1) {Max Pooling};

\node[draw=none, below=0.2cm of pool1] (repeat1) {\footnotesize $\times$6};

\node[block, below=0.7cm of repeat1] (spatialconv) {Conv ($12 \times 1$)};
\node[block, below=0.2cm of spatialconv] (spbn) {Batch Normalization};
\node[block, below=0.2cm of spbn] (sprelu) {ReLU};

\node[block, below=0.7cm of sprelu] (fc1) {Fully Connected};
\node[block, below=0.2cm of fc1] (fcbn1) {Batch Normalization};
\node[block, below=0.2cm of fcbn1] (fcrelu1) {ReLU};
\node[block, below=0.2cm of fcrelu1] (dropout) {Dropout};

\node[draw=none, below=0.2cm of dropout] (repeat2) {\footnotesize $\times$2};

\node[block, below=0.7cm of repeat2, fill=blue!40] (output) {Output (Softmax)};

\draw[connector] (input) -- (conv1);
\draw[connector] (conv1) -- (bn1);
\draw[connector] (bn1) -- (relu1);
\draw[connector] (relu1) -- (pool1);
\draw[connector] (pool1) -- (repeat1);
\draw[connector] (repeat1) -- (spatialconv);
\draw[connector] (spatialconv) -- (spbn);
\draw[connector] (spbn) -- (sprelu);
\draw[connector] (sprelu) -- (fc1);
\draw[connector] (fc1) -- (fcbn1);
\draw[connector] (fcbn1) -- (fcrelu1);
\draw[connector] (fcrelu1) -- (dropout);
\draw[connector] (dropout) -- (repeat2);
\draw[connector] (repeat2) -- (output);

\node[label, right=0.20cm of bn1] (temporal) {
\textbf{Six Layers \\ ``Temporal" Analysis} \\
N Filters shaped as $1 \times K$ \\
$K = 5,5,5,3,3,3$ \\
$N = 16,16,32,32,64,64$ \\
MP factor: $2,2,4,2,2,4$
};

\node[label, right=0.2cm of spbn] (spatial) {
\textbf{Single Layer\\``Spatial" Analysis} \\
Filters shaped as $12 \times 1$
};

\node[label, right=0.2cm of fcbn1] (fcblock) {
\textbf{Two Layers \\ Fully Connected} \\
Each followed by BN,\\ ReLU, Dropout
};

\end{tikzpicture}
\caption{AttiaNet architecture~\cite{attia2019screening}}
\label{fig2:attianet}

\end{figure}

AttiaNet is a compact CNN originally developed for detecting cardiac contractile dysfunction~\cite{attia2019screening}. The model takes as input a \(12 \times T\) ECG matrix (e.g., \(T=5000\) represents a 10-second recording sampled at 500 Hz). As illustrated in Fig.~\ref{fig2:attianet}, the architecture begins with six temporal convolutional blocks. Each block consists of a 1D convolutional layer with kernel size \(1 \times K\), followed by batch normalization, ReLU activation, and max pooling. Kernel sizes decrease (\(K = 5, 5, 5, 3, 3, 3\)) as filter counts increase (\(N = 16, 16, 32, 32, 64, 64\)). Max pooling with stride 2 progressively reduces temporal resolution. After temporal feature extraction, a single spatial convolutional layer with kernel size \(12 \times 1\) integrates information across leads. The resulting representation is fed into two fully connected layers, each followed by batch normalization, ReLU, and dropout. We adopt this architecture as a lightweight baseline, and designate it \textbf{AttiaNet} in reference to the lead author of the original work.

\subsection{ParallelCNN}
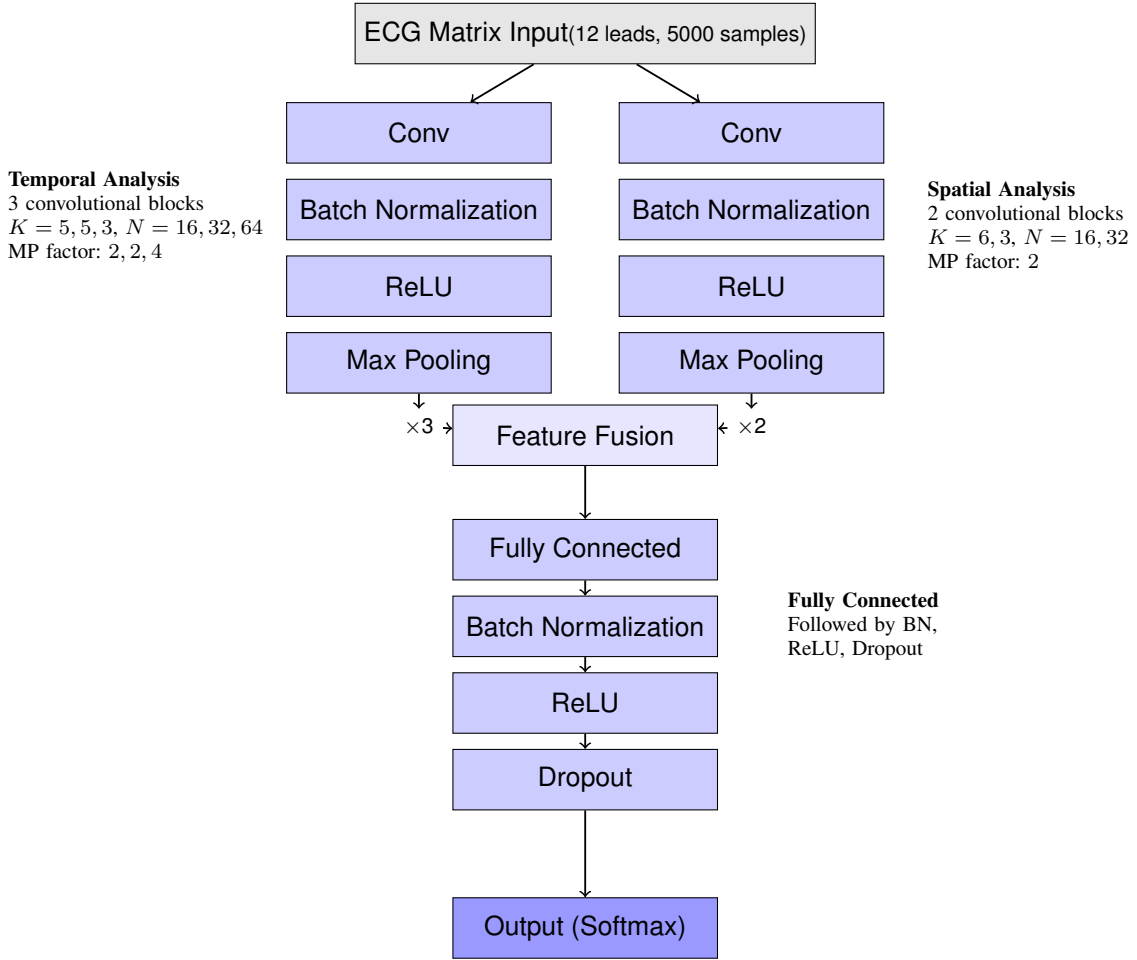
\begin{figure*}[ht]
\centering
\begin{minipage}{0.85\textwidth}  
\begin{tikzpicture}[
    block/.style={rectangle, draw=black, fill=blue!20, minimum width=3.5cm, minimum height=0.8cm, text centered},
    connector/.style={->, line width=0.7pt},
    label/.style={text width=5.5cm, align=left, font=\footnotesize},
    font=\sffamily
]

\node[block, fill=gray!20] (input) {ECG Matrix Input\\ \footnotesize (12 leads, 5000 samples)};

\node[block, below=0.5cm of input, xshift=-2.2cm] (tconv1) {Conv};
\node[block, below=0.2cm of tconv1] (tbn1) {Batch Normalization};
\node[block, below=0.2cm of tbn1] (trelu1) {ReLU};
\node[block, below=0.2cm of trelu1] (tmp1) {Max Pooling};
\node[draw=none, below=0.2cm of tmp1] (tstacknote) {\footnotesize $\times$3};

\node[block, below=0.5cm of input, xshift=2.2cm] (sconv1) {Conv};
\node[block, below=0.2cm of sconv1] (sbn1) {Batch Normalization};
\node[block, below=0.2cm of sbn1] (srelu1) {ReLU};
\node[block, below=0.2cm of srelu1] (smp1) {Max Pooling};
\node[draw=none, below=0.2cm of smp1] (sstacknote) {\footnotesize $\times$2};

\node[block, below=4.5cm of input, fill=blue!10] (fusion) {Feature Fusion};

\node[block, below=0.7cm of fusion] (fc1) {Fully Connected};
\node[block, below=0.2cm of fc1] (fcbn1) {Batch Normalization};
\node[block, below=0.2cm of fcbn1] (fcrelu1) {ReLU};
\node[block, below=0.2cm of fcrelu1] (dropout) {Dropout};
\node[draw=none, below=0.2cm of dropout] (fcstacknote) {};

\node[block, below=0.7cm of fcstacknote, fill=blue!40] (output) {Output (Softmax)};

\draw[connector] (input) -- (tconv1);
\draw[connector] (input) -- (sconv1);
\draw[connector] (tmp1) -- (tstacknote);
\draw[connector] (smp1) -- (sstacknote);
\draw[connector] (tstacknote) -- (fusion);
\draw[connector] (sstacknote) -- (fusion);
\draw[connector] (fusion) -- (fc1);
\draw[connector] (fc1) -- (fcbn1);
\draw[connector] (fcbn1) -- (fcrelu1);
\draw[connector] (fcrelu1) -- (dropout);
\draw[connector] (dropout) -- (output);

\node[label, anchor=west] at ([xshift=-3.8cm,yshift=-0.1cm]tbn1.west) (tannot) {
\textbf{Temporal Analysis} \\
3 convolutional blocks \\
$K = 5, 5, 3$, $N = 16, 32, 64$ \\
MP factor: $2, 2, 4$
};

\node[label, right=0.45cm of sbn1, yshift=-0.2cm, anchor=west] (sannot) {
\textbf{Spatial Analysis} \\
2 convolutional blocks \\
$K = 6, 3$, $N = 16, 32$ \\
MP factor: $2$
};

\node[label, right=0.8cm of fcbn1] (fcannot) {
\textbf{Fully Connected} \\
Followed by BN,\\ ReLU, Dropout
};

\node[draw=none, minimum width=5.5cm] at ([xshift=3.8cm]sbn1.east) {};
\end{tikzpicture}
\end{minipage}
\caption{ParallelCNN architecture: dual temporal and spatial pathways with parallel convolutional blocks and feature fusion.}
\label{fig:parallelcnn}
\end{figure*}

ParallelCNN is a lightweight dual-branch architecture, designed to capture temporal and spatial ECG characteristics explicitly (Fig.~\ref{fig:parallelcnn}). The model consists of three components: (i) Temporal pathway: three convolutional layers with kernels of size \(1 \times K\) (\(K = 5, 5, 3\)), using 16, 32 and 64 filters, respectively. Each layer is followed by batch normalization, ReLU activation, and max pooling with pool sizes \(2, 2, 4\). (ii) Spatial pathway: a convolutional layer with kernel size \(12 \times 1\) captures inter-lead dependencies, followed by batch normalization and ReLU. (iii) Fusion and classification: the outputs of both branches are flattened and concatenated, optionally with demographic information (age and sex). The fused vector is passed through a fully connected layer before the classification layer. The explicit separation of temporal and spatial processing enables efficient yet accurate representation learning.

\subsection{ParallelCNNew}
\label{subsec:parallelcnnew}
ParallelCNNew is a variant of ParallelCNN with an emphasis on balanced learning across branches. Specifically, the temporal and spatial pathways are initialized with equal weights to prevent early-stage bias toward one modality. This design allows the model to adaptively learn the relative importance of temporal versus spatial features during training. Structurally, it shares the same dual-branch design, convolutional modules, pooling strategy, and fusion mechanism as ParallelCNN, differing only in its initialization scheme. The modification aims to improve robustness and stability during training.

\subsection{SimpleNet}
\label{subsec:simplenet}
SimpleNet is a compact, single-stream CNN composed of six sequential convolutional layers followed by two fully connected layers. Unlike the parallel architectures, SimpleNet employs square-shaped kernels to jointly capture spatial (lead) and temporal (time) dependencies within a unified pathway. Each convolutional block is followed by batch normalization, ReLU activation, and max pooling. After the convolutional stack, features are flattened and passed through dense layers, with optional integration of demographic metadata. Owing to its simplicity and small parameter footprint, SimpleNet provides a strong efficiency-focused baseline against which more complex architectures can be compared.

\section{Datasets and Tasks}
\label{sec:datasets}

To evaluate the performance and generalizability of the five CNN architectures---including the three lightweight models we designed---we conducted experiments on three publicly available 12-lead ECG datasets from Germany, China, and the United States. These datasets differ in sample size, patient demographics, labeling granularity, and clinical diversity, thereby providing a robust testbed for assessing diagnostic performance across heterogeneous populations. Each dataset was mapped to a distinct diagnostic task: (i) \textbf{Binary classification} --- distinguishing normal from abnormal ECGs, (ii) \textbf{Multiclass classification} --- identifying mutually exclusive cardiac rhythm categories, and (iii) \textbf{Multilabel classification} --- detecting multiple co-occurring cardiac conditions within a single ECG.

To further examine the role of auxiliary patient information, we evaluated each model under two input settings: (i) raw ECG signals only and (ii) ECG signals integrated with demographic features of the patient (age and sex). These features are routinely collected, computationally inexpensive to incorporate, and clinically relevant, making them an important variable for assessing real-world applicability.

In the following, we describe each dataset and its associated classification task.

\subsection{PTB-XL Dataset}
\label{sec:datasets:ptb-xl}
The Physikalisch-Technische Bundesanstalt (PTB-XL) dataset contains 21,799 10-second, 12-lead clinical ECGs from 18,869 patients, recorded at 500 Hz~\cite{wagner2022ptb,wagner2020ptbp}. The patient cohort is approximately balanced by sex (52\% male, 48\% female) and spans a wide age range (0--95 years; median 62, IQR 22). The ECGs were collected in Germany between 1989 and 1996 using Schiller AG devices and annotated independently by up to two cardiologists. Diagnoses are organized into five superclasses with 24 subclasses: (i) myocardial infarction (MI), (ii) ST/T changes, (iii) conduction disturbances (CD), (iv) hypertrophy (HYP), and (v) normal (NORM). Since each ECG can have multiple labels, PTB-XL constitutes a \textbf{multilabel classification} task in our study.

\subsection{Chapman-Shaoxing Dataset}
\label{sec:datasets:chapman}
The Chapman-Shaoxing dataset includes 12-lead ECGs from 10,646 patients recorded at 500 Hz in Shaoxing People's Hospital, China, using the GE MUSE system~\cite{zheng202012}. The cohort comprises 56\% male and 44\% female patients. Each ECG was annotated with 11 rhythm types and 67 diagnostic codes by expert cardiologists.

For this dataset, we formulated a \textbf{multiclass classification} task by consolidating the 11 rhythm types into four clinically coherent groups:
(i) Atrial Fibrillation (AFIB) --- atrial fibrillation and atrial flutter,(ii) General Supraventricular Tachycardia (GSVT) --- supraventricular tachycardia, atrial tachycardia, AV node re-entrant tachycardia, AV re-entrant tachycardia, and sinus-to-atrial wandering rhythms,(iii) Sinus Bradycardia (SB) --- sinus bradycardia only, and (iv) Sinus Rhythm (SR) --- sinus rhythm and sinus irregularity.

This grouping captures a broad spectrum of clinically relevant rhythm abnormalities and follows established guidelines~\cite{january20192019, page20162015, kirchhof2016esc}.

\subsection{MIMIC-IV-ECG Dataset}
\label{sec:datasets:mimic}
The MIMIC-IV-ECG dataset contains approximately 800,000 diagnostic 12-lead ECGs from nearly 160,000 patients admitted to Beth Israel Deaconess Medical Center (Boston, USA) between 2008 and 2019~\cite{goldberger2000physiobank, gow2023mimic}. All recordings are 10 seconds in duration, sampled at 500 Hz, and linked to corresponding cardiologist reports. The cohort consists of 53\% female and 47\% male patients.

We formulated a \textbf{binary classification} task by distinguishing normal from abnormal ECGs. Specifically, ECGs labeled as ``sinus rhythm'' were categorized as normal, while all others with any documented abnormality were categorized as abnormal.

\section{Experiments and Results}
\label{experiments_and_results}
We evaluated the five CNN architectures across three diagnostic tasks---binary,
multiclass, and multilabel classification---using the datasets described in
Section~\ref{sec:datasets}. All models were trained with the Adam optimizer
(learning rate \(1 \times 10^{-3}\), batch size 32), with early stopping based
on validation loss to mitigate overfitting. Training and inference were performed on a single NVIDIA GeForce RTX 3070 GPU (8~GB).

Performance was primarily measured using the Area Under the Receiver Operating
Characteristic Curve (AUC), a widely adopted and clinically relevant metric for
diagnostic classification. To assess model deployability, we further evaluated
computational efficiency in terms of parameter count, inference time, memory
footprint, and overall efficiency score (see Section~\ref{sec:efficiency}).

Finally, to examine the role of patient metadata, we compared each model in two
configurations: (i) raw ECG input only, and (ii) ECG input integrated with
demographic features (age and sex). Figures~\ref{fig:auc_multilabel}--\ref{fig:auc_binary} present AUC scores across tasks under both configurations.
Tables~\ref{tab:multilabel}--\ref{tab:binary} report detailed efficiency results across the three classification tasks.

\begin{figure}[ht]
\centering
\begin{tikzpicture}
\begin{axis}[
    ybar, bar width=8pt, width=8.5cm, height=5cm, axis lines=left,
    ymin=0, ymax=1.0, enlarge x limits=0.15,
    symbolic x coords={1,2,3,4,5}, xtick=data,
    xticklabels={DeepResidualCNN, AttiaNet, ParallelCNN, ParallelCNNew, SimpleNet},
    x tick label style={rotate=45, anchor=east},
    ylabel={\textbf{AUC}}, ymajorgrids=true, grid style={dashed, gray!40},
    legend style={at={(0.5,1.3)}, anchor=north, legend columns=2}
]
\addplot coordinates {(1,0.64) (2,0.99) (3,0.99) (4,0.99) (5,0.99)};
\addplot coordinates {(1,0.71) (2,0.99) (3,0.99) (4,0.99) (5,0.99)};
\legend{Without Age \& Sex, With Age \& Sex}
\end{axis}
\end{tikzpicture}

\caption{AUC scores for multilabel classification.}
\label{fig:auc_multilabel}
\end{figure}
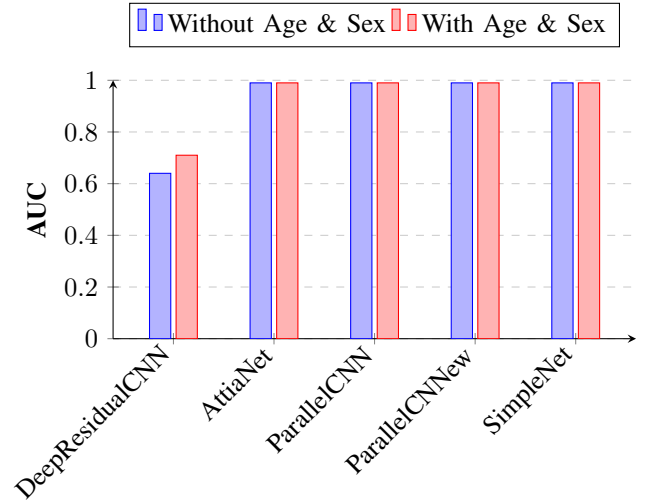

\begin{figure}[ht]
\centering
\begin{tikzpicture}
\begin{axis}[
    ybar, bar width=8pt, width=8.5cm, height=5cm, axis lines=left,
    ymin=0, ymax=1.0, enlarge x limits=0.15,
    symbolic x coords={1,2,3,4,5}, xtick=data,
    xticklabels={DeepResidualCNN, AttiaNet, ParallelCNN, ParallelCNNew, SimpleNet},
    x tick label style={rotate=45, anchor=east},
    ylabel={\textbf{AUC}}, ymajorgrids=true, grid style={dashed, gray!40},
    legend style={at={(0.5,1.3)}, anchor=north, legend columns=2}
]
\addplot coordinates {(1,0.61) (2,0.99) (3,0.97) (4,0.97) (5,0.97)};
\addplot coordinates {(1,0.66) (2,0.99) (3,0.98) (4,0.98) (5,0.97)};
\legend{Without Age \& Sex, With Age \& Sex}
\end{axis}
\end{tikzpicture}

\caption{AUC scores for multiclass classification.}
\label{fig:auc_multiclass}
\end{figure}
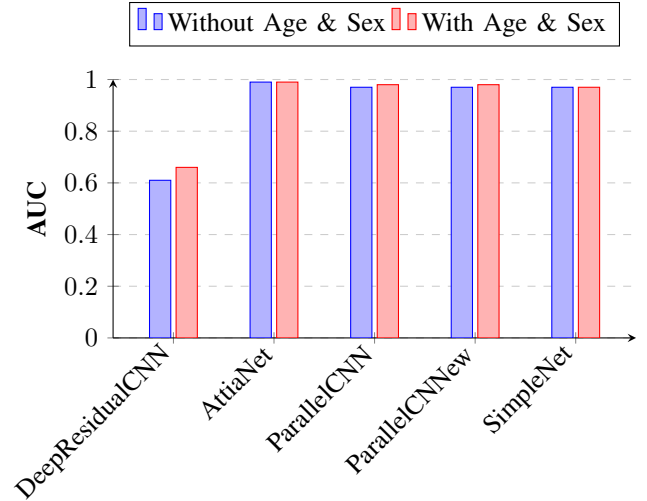

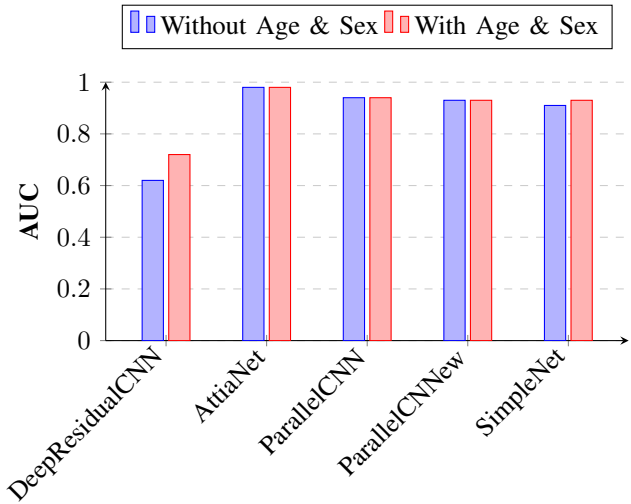
\begin{figure}[ht]
\centering
\begin{tikzpicture}
\begin{axis}[
    ybar, bar width=8pt, width=8.5cm, height=5cm, axis lines=left,
    ymin=0, ymax=1.0, enlarge x limits=0.15,
    symbolic x coords={1,2,3,4,5}, xtick=data,
    xticklabels={DeepResidualCNN, AttiaNet, ParallelCNN, ParallelCNNew, SimpleNet},
    x tick label style={rotate=45, anchor=east},
    ylabel={\textbf{AUC}}, ymajorgrids=true, grid style={dashed, gray!40},
    legend style={at={(0.5,1.3)}, anchor=north, legend columns=-1}
]
\addplot coordinates {(1,0.62) (2,0.98) (3,0.94) (4,0.93) (5,0.91)};
\addplot coordinates {(1,0.72) (2,0.98) (3,0.94) (4,0.93) (5,0.93)};
\legend{Without Age \& Sex, With Age \& Sex}
\end{axis}
\end{tikzpicture}
\caption{AUC scores for binary classification.}
\label{fig:auc_binary}
\end{figure}

\begin{table*}[ht]
\centering

\caption{Efficiency Comparison: Multilabel Classification Task}
\begin{tabular}{lrrrrr}
\toprule
\textbf{Model} & \textbf{Params (in millions)} & \textbf{Inference Time (ms)} & \textbf{Peak GPU Memory (MB)} & \textbf{Resource Cost} & \textbf{Efficiency Score} \\
\midrule
AttiaNet          & \textbf{0.15} & 0.79 & \textbf{48.37} & \textbf{0.15} & \textbf{0.93} \\
SimpleNet         & \underline{2.59} & \underline{0.62} & 244.43 & \underline{0.42} & \underline{0.82} \\
ParallelCNN       & 20.42 & \textbf{0.51} & \underline{149.06} & 0.47 & 0.81 \\
ParallelCNNew     & 22.99 & 0.63 & 236.67 & 0.72 & 0.71 \\
DeepResidualCNN   & 5.90  & 1.16 & \underline{77.17}  & 0.47 & 0.59 \\
\bottomrule
\end{tabular}
\label{tab:multilabel}
\end{table*}

\begin{table*}[ht]
\centering
\caption{Efficiency Comparison: Multiclass Classification Task}
\begin{tabular}{lrrrrr}
\toprule
\textbf{Model} & \textbf{Params (in millions)} & \textbf{Inference Time (ms)} & \textbf{Peak GPU Memory (MB)} & \textbf{Resource Cost} & \textbf{Efficiency Score} \\
\midrule
AttiaNet          & \textbf{0.15} & 0.86 & \textbf{48.37} & \textbf{0.17} & \textbf{0.93} \\
SimpleNet         & \underline{2.59} & \textbf{0.39} & 244.42 & \underline{0.37} & \underline{0.83} \\
ParallelCNN       & 20.42 & 0.99 & 149.06 & 0.51 & 0.71 \\
ParallelCNNew     & 22.99 & \underline{0.64} & 236.67 & 0.74 & 0.68 \\
DeepResidualCNN   & 5.90  & 1.32 & \underline{77.17}  & 0.47 & 0.58 \\
\bottomrule
\end{tabular}
\label{tab:multiclass}
\end{table*}

\begin{table*}[ht]
\centering
\caption{Efficiency Comparison: Binary Classification Task}
\begin{tabular}{lrrrrr}
\toprule
\textbf{Model} & \textbf{Params (in millions)} & \textbf{Inference Time (ms)} & \textbf{Peak GPU Memory (MB)} & \textbf{Resource Cost} & \textbf{Efficiency Score} \\
\midrule
AttiaNet          & \textbf{0.15} & 0.75 & \textbf{48.37} & \textbf{0.18} & \textbf{0.91} \\
SimpleNet         & \underline{2.59} & \textbf{0.34} & 244.42 & \underline{0.37} & \underline{0.80} \\
ParallelCNN       & 20.41 & \underline{0.45} & 149.05 & 0.52 & 0.76 \\
ParallelCNNew     & 22.98 & 0.54 & 236.66 & 0.75 & 0.66 \\
DeepResidualCNN   & 5.90  & 1.08 & \underline{77.16}  & 0.47 & 0.59 \\
\bottomrule
\end{tabular}
\label{tab:binary}
\end{table*}

\subsection{Main Results}
Across all tasks, several key findings emerged:
(i) AttiaNet achieved consistently high diagnostic accuracy, with AUC scores near 0.99 in the multilabel and multiclass tasks and 0.98 in the binary task.

(ii) Our three lightweight models (ParallelCNN, ParallelCNNew, and SimpleNet) delivered competitive performance, especially in the multilabel task, where all three achieved AUC scores of 0.99. In the multiclass and binary settings, their performance remained high, though slightly below AttiaNet. This small gap may reflect the use of fixed hyperparameters across models rather than task-specific tuning.
(iii) DeepResidualCNN performed worst across all tasks, with AUCs of 0.64 (multilabel), 0.61 (multiclass), and 0.62 (binary). Despite its depth and complexity, the model underperformed relative to compact alternatives, likely due to overparameterization and insufficient regularization.

Overall, these results suggest that well-designed compact architectures can achieve state-of-the-art accuracy while avoiding the inefficiencies of complex models, making them more suitable for deployment in real-time clinical workflows.

\subsection{Incorporating Demographic Features (Age and Sex)}
Including demographic information produced modest but measurable improvements across most models. The most pronounced gains occurred in DeepResidualCNN, with increases in AUC of 7\%, 5\%, and 10\% for multilabel, multiclass, and binary tasks, respectively. In contrast, AttiaNet, SimpleNet, ParallelCNN, and ParallelCNNew showed only marginal improvements.

These findings suggest that lightweight architectures already capture most of the discriminatory signals from ECG waveforms, leaving limited additional benefit from demographic features. Nevertheless, the consistent (albeit small) improvements highlight the potential value of integrating inexpensive metadata in clinical AI systems, particularly for models that struggle with signal-only learning.

\subsection{Computational Efficiency Analysis}
\label{sec:efficiency}
We next assessed computational efficiency using three metrics: parameter count (in millions), average inference time per ECG (ms), and peak GPU memory usage (MB). To integrate these factors, we defined a composite \textit{Resource Cost} as the mean of min--max normalized values of parameters, inference time, and memory usage:

\begin{equation}
\text{ResourceCost} = \frac{1}{3} (P + T + M),
\end{equation}

where \(P\), \(T\), and \(M\) are normalized values for parameter count, inference time, and memory usage.

To balance diagnostic accuracy with computational efficiency, we propose a unified evaluation metric, termed the \textit{Efficiency Score}.

\begin{equation}
\text{EfficiencyScore} = \lambda \cdot \text{AUC} + (1 - \lambda) \cdot (1 - \text{ResourceCost}),
\end{equation}

where \(\lambda = 0.6\), placing slightly greater emphasis on classification performance while still rewarding computational efficiency. In practice, this means that models with high diagnostic accuracy but excessive computational cost are penalized, while lightweight models that achieve competitive accuracy can still obtain strong Efficiency Scores---reflecting the trade-offs necessary for real-time clinical deployment.

Across all tasks, AttiaNet achieved the highest overall efficiency, scoring 0.93 on both multilabel and multiclass tasks and 0.91 on the binary task, owing to its compact design and strong diagnostic accuracy. SimpleNet consistently ranked second, but outperformed AttiaNet in inference
speed across all tasks, highlighting its suitability for time-critical
applications. ParallelCNN and ParallelCNNew provided a balance between speed and accuracy, though their relatively larger parameter counts reduced efficiency scores. By contrast, DeepResidualCNN was the least efficient, demonstrating that increased architectural depth does not necessarily yield better clinical utility.

In summary, AttiaNet offers the best overall balance of accuracy and efficiency, while SimpleNet stands out as particularly attractive for real-time monitoring and edge-device deployment due to its superior inference speed. ParallelCNN and ParallelCNNew illustrate the potential of dual-branch spatiotemporal designs, albeit with slightly added complexity. DeepResidualCNN, despite its prior benchmark success, appears less suitable for resource-constrained clinical environments. Taken together, these findings support the central argument of this study: carefully designed, lightweight CNN architectures can achieve near state-of-the-art diagnostic accuracy while enabling real-time ECG interpretation on resource-constrained hardware, thereby enhancing overall efficiency.

\section{Discussion}
\label{discussion}
This study offers a practical alternative to recent trends in ECG analysis that prioritize large foundational models with substantial computational requirements~\cite{vaid2023foundational, mckeen2024ecg}. While such models achieve strong performance, their size and training demands limit real-world deployment. In contrast, we demonstrate that lightweight, task-specific CNNs---such as SimpleNet and AttiaNet---deliver competitive diagnostic accuracy across binary, multiclass, and multilabel tasks while significantly reducing inference time, parameter count, and memory usage.

A key contribution of this work is the introduction of an \textit{Efficiency Score}, a unified metric that jointly balances diagnostic performance and computational cost, enabling deployment-aware benchmarking for fair model comparison. We also explored the role of demographic metadata (age and sex), finding that while these low-cost features improved performance in deeper models like DeepResidualCNN, they provided minimal benefit to compact architectures that already capture dominant waveform patterns. This underscores a nuanced role for auxiliary patient features, particularly in resource-constrained environments.

Clinically, our models show strong potential for real-time use in mobile or point-of-care ECG devices. AttiaNet offers the best trade-off between accuracy and efficiency, whereas SimpleNet achieves superior inference speed for edge deployment. Their consistent performance across three large, publicly available
datasets from Germany, China, and the United States further suggests generalizability across diverse populations.

However, the limitations of our work include the use of fixed training hyperparameters, reliance on static ECG snapshots rather than continuous monitoring, and efficiency benchmarking conducted on a mid-range GPU (NVIDIA RTX 3070, 8GB). While this provides a conservative and realistic estimate of deployability compared to high-end server GPUs, future work should include direct benchmarking on mobile and embedded hardware. Additional directions include adaptive training strategies, richer multimodal inputs, and prospective validation in clinical workflows to confirm practical utility.

\section{Conclusion}
\label{conclusion}
In this study, we systematically evaluated five CNN architectures for efficient and accurate ECG interpretation, including our three proposed lightweight models---ParallelCNN, ParallelCNNew, and SimpleNet---and two established baselines: AttiaNet and DeepResidualCNN. Using three diverse 12-lead ECG datasets and multiple classification tasks (binary, multiclass, and multilabel), we showed that carefully designed, task-specific lightweight architectures can achieve near state-of-the-art diagnostic accuracy while substantially reducing computational costs. To enable fair benchmarking, we introduced an \textit{Efficiency Score} that integrates model size, inference speed, memory usage, and diagnostic performance.

Among the evaluated models, AttiaNet achieved the highest overall efficiency, offering the best balance between accuracy and resource use, while SimpleNet delivered superior inference speed with only a marginal drop in performance---making it particularly suitable for real-time and edge-device deployment. ParallelCNN and ParallelCNNew further demonstrated the viability of dual-branch spatiotemporal designs, albeit with increased complexity. Overall, these findings highlight that compact CNNs such as AttiaNet and SimpleNet represent practical, sustainable solutions for real-time ECG interpretation in both advanced and resource-limited healthcare settings.

\bibliographystyle{ieeetr}
\bibliography{references}

\end{document}